\documentclass[lettersize,journal]{IEEEtran}
\usepackage{amsmath,amsfonts}
\usepackage{algorithmic}
\usepackage{algorithm}
\usepackage{array}
\usepackage[caption=false,font=normalsize,labelfont=sf,textfont=sf]{subfig}
\usepackage{textcomp}
\usepackage{stfloats}
\usepackage{url}
\usepackage{verbatim}
\usepackage{graphicx}
\usepackage{cite}
\usepackage{multirow}
\usepackage{diagbox}
\usepackage{bbding}
\usepackage{booktabs}
\usepackage{makecell}
\usepackage{color}
\usepackage{booktabs}       
\usepackage[center]{caption}
\begin{document}

\title{SpeechColab Leaderboard: An Open-Source Platform for Automatic Speech Recognition Evaluation}

\author{Jiayu~Du,  Jinpeng~Li, Guoguo~Chen, and Wei-Qiang~Zhang,~\IEEEmembership{Senior~Member,~IEEE}
\thanks{J. Du and J. Li contributed equally to this work and should be considered co-first author.} 
\thanks{Jiayu Du, SpeechColab, China (email: jerry.jiayu.du@gmail.com).}
\thanks{Jinpeng Li, Department of Electrical Engineering, Tsinghua University, Beijing 100084, China (email: lijp22@mails.tsinghua.edu.cn).}
\thanks{Guoguo Chen, Seasalt AI Inc., USA; SpeechColab, China (email:  chenguoguo06@gmail.com).}
\thanks{Wei-Qiang Zhang, Department of Electrical Engineering, Tsinghua University, Beijing 100084, China; SpeechColab, China (email:  wqzhang@tsinghua.edu.cn).}
\thanks{This work was supported in part by the National Natural Science Foundation of China under Grant 62276153.}}



\maketitle

\begin{abstract}
In the wake of the surging tide of deep learning over the past decade, Automatic Speech Recognition (ASR) has garnered substantial attention, leading to the emergence of numerous publicly accessible ASR systems that are actively being integrated into our daily lives. Nonetheless, the impartial and replicable evaluation of these ASR systems encounters challenges due to various crucial subtleties. In this paper we introduce the SpeechColab Leaderboard, a general-purpose, open-source platform designed for ASR evaluation. With this platform:
(i) We report a comprehensive benchmark, unveiling the current state-of-the-art panorama for ASR systems, covering both open-source models and industrial commercial services.
(ii) We quantize how distinct nuances in the scoring pipeline influence the final benchmark outcomes. These include nuances related to capitalization, punctuation, interjection, contraction, synonym usage, compound words, etc. These issues have gained prominence in the context of the transition towards an End-to-End future.
(iii) We propose a practical modification to the conventional Token-Error-Rate (TER) evaluation metric, with inspirations from Kolmogorov complexity and Normalized Information Distance (NID). This adaptation, called modified-TER (mTER), achieves proper normalization and symmetrical treatment of reference and hypothesis.
By leveraging this platform as a large-scale testing ground, this study demonstrates the robustness and backward compatibility of mTER when compared to TER. 
The SpeechColab Leaderboard is accessible at \url{https://github.com/SpeechColab/Leaderboard}.
\end{abstract}


\section{Introduction}
Automatic Speech Recognition (ASR) has been an active research topic for many years.
Traditional ASR combines Hidden Markov Models (HMM) and Gaussian Mixture Models (GMM) to capture the dynamics of speech signal and the hierarchical knowledge behind human languages \cite{deng1991phonemic}.
In recent years, deep neural networks (DNN) have started to emerge with superior accuracy \cite{dahl2011context}, and have quickly become the mainstream for ASR.
For instance,
chain model \cite{povey2016purely} incorporates Convolutional Neural Networks (CNNs) and Time Delay Neural Network (TDNNs),
while DeepSpeech \cite{hannun2014deep} model utilizes Recurrent Neural Networks (RNNs) and Long Short-Term Memory Networks (LSTMs).
Modern systems are leaning to even more sophisticated architectures such as Transformer \cite{vaswani2017attention} and Conformer \cite{gulati2020conformer},
coupled with sequence losses like Connectionist Temporal Classification (CTC) \cite{CTC} and Recurrent Neural Network Transducer (RNN-T) \cite{graves2012sequence}.
From a system perspective, driven by the scaling law from language modeling research,
large speech models have been developed such as OpenAI-Whisper \cite{radford2022robust}, and Google-USM \cite{zhang2023google},
pushing up the scale of ASR training by orders of magnitude.
In the meantime, self-supervised training, as a paradigm shift,
is also gaining popularity to leverage abundant unlabeled data in the world.
Notable examples are wav2vec 2.0 \cite{baevski2020wav2vec}, HuBERT \cite{hsu2021hubert}, WavLM \cite{chen2022wavlm}, and data2vec \cite{baevski2022data2vec}.

Given the swift evolution of ASR technology, a variety of speech toolkits have been developed and open-sourced,
such as HTK \cite{young2002htk}, Kaldi \cite{povey2011kaldi}, 
ESPnet \cite{watanabe2018espnet}, NeMo \cite{kuchaiev2019nemo}, 
SpeechBrain \cite{ravanelli2021speechbrain}, WeNet \cite{yao2021wenet}, and K2\footnote{https://github.com/k2-fsa},
offering comprehensive libraries and recipes to facilitate ASR research and development.
\IEEEpubidadjcol
However, the evaluation of ASR still remains challenging \cite{del2021earnings} \cite{gandhi2022esb},
because there exist various crucial subtleties and pitfalls that require non-trivial efforts to do right in practice,
such as text normalization \cite{faria2022toward}.
The divergent ecosystem struggles to reach a clear and consistent understanding on the performance of modern ASR systems. 

To address the problem, we present \emph{SpeechColab Leaderboard}, an open-source benchmark platform,
so that speech researchers and developers can reliably reproduce, examine, and compare all kinds of ASR systems.
The platform is designed to be:
(i) \textit{Simple}: consistent data formats and unified interfaces minimize accidental complexity.
(ii) \textit{Open}: leaderboard users should be able to easily share and exchange resources (e.g. test sets, models, configurations).
(iii) \textit{Reproducible}: ASR systems, including all their dependencies and environment details, should be reproducible as a whole.

In Section 2, we describe the proposed platform, including three major components:
\textit{a dataset zoo}, \textit{a model zoo}, and \textit{an evaluation pipeline}.
In Section 3, we report a large-scale benchmark for English ASR on the platform.
In Section 4, the traditional evaluation metric TER (Token Error Rate) is briefly revisited,
and a simple and practical modification is proposed to make TER more robust.

\begin{figure*}[ht]
    \centering
    \includegraphics[width = 0.95\textwidth]{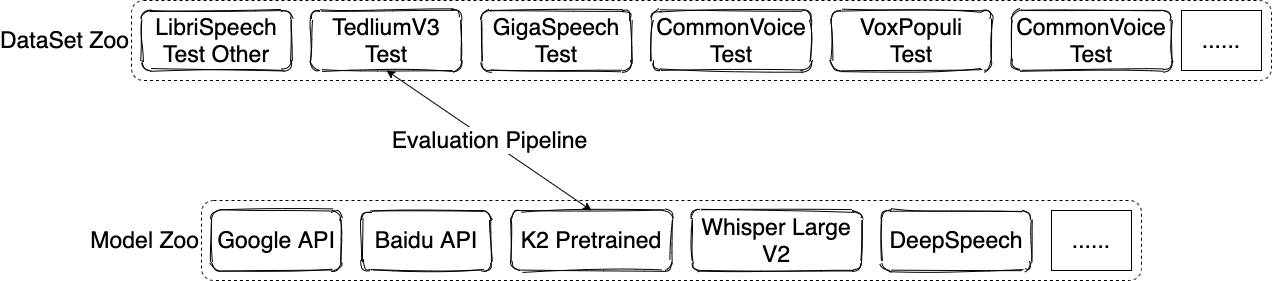}
    \caption{SpeechColab Leaderboard}
    \label{fig:leaderboard_zoo}
\end{figure*}

\begin{table*}[ht]
  \caption{Dataset Zoo}
  \label{tab:dataset}
  \centering
  \scalebox{1}{
  \begin{tabular}{cccccc}
    \toprule
    \textbf{Dataset}  & \textbf{\makecell{Number of\\ sentences}} & \textbf{Total duration} & \textbf{Source} & \textbf{Style} & \textbf{Release Date}  \\
    \midrule
    LibriSpeech.test-clean \cite{panayotov2015librispeech} & 2620 & 5.403 hours & the LibriVox project & narrated audio-books & 2015\\
    LibriSpeech.test-other \cite{panayotov2015librispeech} & 2939 & 5.342 hours & the LibriVox project & narrated audio-books & 2015 \\
    TEDLIUM3.dev \cite{hernandez2018ted} & 507 & 1.598 hours & TED talks & oratory & 2018 \\
    TEDLIUM3.test \cite{hernandez2018ted} & 1155 & 2.617 hours & TED talks & oratory & 2018 \\
    GigaSpeech.dev \cite{chen2021gigaspeech} & 5715 & 11.366 hours & Podcast and YouTube & spontaneous & 2021 \\
    GigaSpeech.test \cite{chen2021gigaspeech} & 19930 & 35.358 hours & Podcast and YouTube & spontaneous & 2021 \\
    VoxPopuli.dev \cite{wang2021voxpopuli} & 1753 & 4.946 hours & European Parliament & spontaneous & 2021 \\
    VoxPopuli.test \cite{wang2021voxpopuli} & 1841 & 4.864 hours & European Parliament & spontaneous & 2021 \\
    VoxPopuli.test-accented \cite{wang2021voxpopuli} & 8357 & 26.174 hours & European Parliament & spontaneous & 2022 \\
    CommonVoice11.0.dev \cite{ardila2020common} & 16352 & 27.245 hours & crowd sourcing & narrated prompts & 2022\\
    CommonVoice11.0.test \cite{ardila2020common} & 16351 & 26.950 hours & crowd sourcing & narrated prompts & 2022 \\
    \bottomrule
  \end{tabular}
  }
\end{table*}

\begin{table*}[ht]
  \caption{Model Zoo}
  \label{tab:models}
  \centering
  \scalebox{1}{
  \begin{tabular}{ccccc}
    \toprule
    \textbf{Type} & \textbf{Model} & \textbf{Architecture}  & \textbf{Size in bytes}  & \textbf{Date of Evaluation} 
    \\
    \midrule
    \multirow{7}{3cm}{\centering Commercial API} & aliyun\_api\_en &\multirow{7}*{\centering --} &\multirow{7}*{\centering --} & 2022.10 \\
    ~ & amazon\_api\_en & ~ & ~  & 2022.10 \\
    ~ & baidu\_api\_en & ~ & ~ & 2022.10 \\
    ~ & google\_api\_en  & ~ & ~  & 2022.10\\
    ~ & google\_USM\_en \cite{zhang2023google} & ~ & ~ & 2023.03 \\
    ~ & microsoft\_sdk\_en & ~ & ~ & 2022.10 \\
    ~ & tencent\_api\_en & ~ & ~ & 2022.10 \\
    \midrule
    \multirow{8}{3cm}{\centering Open-Source (Supervised) } & vosk\_model\_en\_large & Kaldi chain model & 2.7G & 2022.10 \\
    ~ & deepspeech\_model\_en \cite{hannun2014deep} & RNN + N-gram & 1.1G & 2022.10\\
    ~ & coqui\_model\_en  & RNN + N-gram & 979M & 2022.10\\
    ~ & nemo\_conformer\_ctc\_large\_en \cite{kuchaiev2019nemo} & Conformer-CTC & 465M & 2022.10\\
    ~ & nemo\_conformer\_transducer\_xlarge\_en \cite{kuchaiev2019nemo} & Conformer-Transducer & 2.5G & 2022.10\\
    ~ & k2\_gigaspeech \cite{kuang2022pruned}  & Pruned stateless RNN-T & 320M & 2022.10\\
    ~ & whisper\_large\_v1 \cite{radford2022robust} & Transformer Encoder-Decoder & 2.9G & 2022.10 \\
    ~ & whisper\_large\_v2 \cite{radford2022robust} & Transformer Encoder-Decoder & 2.9G   & 2023.03  \\
    \midrule
    \multirow{4}{3cm}{\centering Open-Source (Unsupervise + Fine-tuned)} & data2vec\_audio\_large\_ft\_libri\_960h  & data2vec \cite{baevski2022data2vec} & 1.2G  & 2022.10  \\
    ~ & hubert\_xlarge\_ft\_libri\_960h  & HuBERT \cite{hsu2021hubert} & 3.6G & 2022.10 \\
    ~ & wav2vec2\_large\_robust\_ft\_libri\_960h  & wav2vec 2.0 \cite{baevski2020wav2vec} & 2.4G   & 2022.10 \\
    ~ & wavlm\_base\_plus\_ft\_libri\_clean\_100h & WavLM \cite{chen2022wavlm} & 361M   & 2022.10  \\
    \bottomrule
  \end{tabular}
  }
  \begin{minipage}[c]{0.9\textwidth}
    \vspace{0.1cm}
    * Some models do not provide interfaces to query the total number of parameters, so we list model sizes in bytes in the table for consistency.
  \end{minipage}
\end{table*}

\section{The Platform}
\label{sec:platform}

\subsection{Overview}
The overall platform is shown in Figure \ref{fig:leaderboard_zoo}.
We implement the dataset zoo and the model zoo on top of commercial cloud-storage services,
so that the platform can serve as a reliable and high-speed disk for our users to exchange data and models.
Evaluation sets and models are associated with globally unique identifiers.
Within our repository, to initiate a benchmark:
\begin{verbatim}
ops/benchmark -m <MODEL_ID> -d <DATASET_ID>
\end{verbatim}

\subsection{Dataset Zoo}
\label{sec:dataset_zoo}

In the dataset zoo:
(i) All utterances, if necessary, are extracted from raw long audio,
and stored consistently in short\footnote{The utterances longer than 60s are removed.} WAV format.
(ii) The metadata is organized in tab-separated-values (.tsv) format with 4 columns, as shown in Table \ref{tab:example_tsv}.
\begin{table}[H]
\caption{Unified metadata.tsv }
\label{tab:example_tsv}
\centering
\scalebox{0.81}{
    \begin{tabular}{cccc}
    
    \toprule
    \textbf{ID} & \textbf{AUDIO} & \textbf{DURATION} & \textbf{TEXT} \\
    \midrule
    POD0000051 & audio/POD0000051.wav & 2.100 & But what kind of business? \\
    POD0000094 & audio/POD0000094.wav & 2.727 & So we're gonna make it ... \\
    ... & ... & ... & ... \\
    \bottomrule
    \end{tabular}
}
\end{table}

As of the writing of this paper, 11 well-known ASR evaluation sets are processed and integrated into the dataset zoo,
summarized in Table \ref{tab:dataset}.
We provide simple operational utilities for dataset management.
For example, to share a local test set to the zoo:
\begin{verbatim}
    ops/push -d <DATASET_ID>
\end{verbatim}
And to retrieve a test set from the zoo:
\begin{verbatim}
    ops/pull -d <DATASET_ID>
\end{verbatim}

\subsection{Model Zoo}
\label{sec:model_zoo}

We define a simple and unified interface for model zoo:
all models should be built within a Docker container along with an ASR program that takes a list of WAV files as input.
By design, we support both open-source models and commercial API services.
Similar to the dataset zoo,
leaderboard users can easily publish or reproduce ASR systems by using:
\begin{verbatim}
    ops/push -m <MODEL_ID>
    ops/pull -m <MODEL_ID>
\end{verbatim}
Table \ref{tab:models} provides a detailed list of 19 integrated models.

\subsection{Evaluation Pipeline}
\label{sec:pipeline}

\begin{table*}[hb]
  \caption{Text Normalization Components}
  \label{tab:tn_example}
  \centering
\scalebox{1}{
  \begin{tabular}{cccc}
    \toprule
    \textbf{\makecell{Pipeline \\ components}}  & \textbf{Description} & \textbf{Raw text} & \textbf{New text}   \\
    \midrule
    CASE & unify cases & And then there was Broad Street. & \makecell{AND THEN THERE \\ WAS BROAD STREET.} \\
    \midrule
    PUNC & \makecell{remove punctuations\\(, . ? ! " - and \\ single quote)} & \makecell{\textcolor{red}{"'}He doesn\textcolor{blue}{'}t say exactly what it is\textcolor{red}{,'}\\ said Ruth\textcolor{red}{,} a little dubiously\textcolor{red}{. "}} & \makecell{He doesn\textcolor{blue}{'}t say exactly what it is \\ said Ruth a little dubiously} \\
    \midrule
    ITJ & remove interjections & \makecell{\textcolor{red}{uh} yeah \textcolor{red}{um} that's good } & yeah that's good\\
    \midrule
    UK-US & \makecell{unify UK \& US  \\ spelling conventions} & \makecell{
        1. she went to the \textcolor{red}{theatre} \\
        2. such a \textcolor{red}{humour} \\
        3. I \textcolor{red}{apologise} \\
    } &
    \makecell{
        1. she went to the \textcolor{blue}{theater} \\
        2. such a \textcolor{blue}{humor} \\
        3. I \textcolor{blue}{apologize} \\
    }\\
    \midrule
    NSW & \makecell{normalize Non-Standard-Word \\ (number, quantity, date,\\ time, address etc)} &
    \makecell{
        1. gave him \textcolor{red}{\$100}.\\
        2. Just before \textcolor{red}{8.30 a.m.}\\
        3. grew up in the \textcolor{red}{1980s} \\
        4. the baggage is \textcolor{red}{12.7kg} \\
        5. in the \textcolor{red}{21st} century \\
        6. \textcolor{red}{1/3} of the population \\
        7. \textcolor{red}{13,000} people \\
        8. \textcolor{red}{1998/2/30}
    } & 
    \makecell{
        1. gave him \textcolor{blue}{one hundred dollars}. \\
        2. Just before \textcolor{blue}{eight thirty AM} \\
        3. grew up in the \textcolor{blue}{nineteen eighties} \\
        4. the baggage is \textcolor{blue}{twelve point seven kilograms} \\
        5. in the \textcolor{blue}{twenty first} century \\
        6. \textcolor{blue}{one third} of the population \\
        7. \textcolor{blue}{thirteen thousand} people \\
        8. \textcolor{blue}{february thirtieth nineteen ninety eight} \\
    } \\
    \bottomrule
  \end{tabular}
  }
\end{table*}

\subsubsection{Preprocessing}
In practice, references and hypotheses are usually in different forms, therefore pre-processing is needed to remove the discrepancy.
As listed in table \ref{tab:tn_example}, we deal with case,
punctuation, common interjections, US/UK spelling convention\footnote{http://www.tysto.com/uk-us-spelling-list.html} etc.
For Non-Standard-Word (NSW) normalization, we leverage the context-dependent rewriting rules from NeMo toolkit \cite{bakhturina2022shallow}.
\footnote{We currently do not have reliable solution for stutter and dis-fluency detection, this could be one of our future works.}


\begin{table*}[ht]
  \caption{Dynamic Alternative Expansion(DAE) (hypothesis only) }
  \label{tab:DAE_example}
  \centering
\scalebox{0.9}{
  \begin{tabular}{ccccc}
    \toprule
    \textbf{\makecell{Pipeline \\ components}} & \textbf{Description} & \textbf{Alternative Set} & \textbf{Raw hyp} & \textbf{hyp after DAE}   \\
    \midrule
    DAE &
    \makecell{contractions, \\ abbreviations, \\compounds \\ etc... } & 
    \makecell{
        We're = We are \\
        I'm = I am, \\ gonna = going to, \\ OK = O K = Okey \\
        storyteller = story-teller \\ = story teller 
    } &
    \makecell{
        1.\textcolor{red}{We're} here early \\
        2.\textcolor{red}{I{'}m gonna} be \textcolor{red}{OK} \\
        3.He is an excellent \textcolor{red}{storyteller} 
    } & 
    \makecell{
        1.(\textcolor{blue}{We{'}re\textbar We are}) here  early \\
        2.(\textcolor{blue}{I'm\textbar I am}) (\textcolor{blue}{gonna\textbar going to}) be (\textcolor{blue}{OK\textbar O K\textbar Okay}) \\
        3.He is an excellent (\textcolor{blue}{storyteller\textbar story teller\textbar story-teller})
    } \\
    \bottomrule
  \end{tabular}
  }
\end{table*}

\subsubsection{Metric}
Token Error Rate is used to evaluate the accuracy of ASR systems:
\begin{equation}
\begin{split}
    {\rm TER}(ref, hyp) & \doteq \frac{\mathcal{LD}(ref \to hyp)}{ |ref| } \label{eq:vanilla-TER}
\end{split}
\end{equation}
where $\mathcal{LD}$ denotes the Levenshtein Distance,
and $|ref|$ refers to the number of words (for English ASR) in the reference.

\subsubsection{Scoring}
The scoring module is implemented in Weighted Finite State Transducer (WFST) framework,
in particular via \emph{OpenFST} \cite{allauzen2007openfst} and \emph{Pynini} \cite{gorman2016pynini}.
The pre-processed references and hypotheses are first tokenized as word sequences,
then transformed to linear FSTs.
The Levenshtein transducer $\mathcal{L}$ is constructed with the standard operations and costs: \textit{insertion} (INS:1.0), \textit{deletion} (DEL:1.0), \textit{substitution} (SUB:1.0), and \textit{correct-match} (COR:0.0).
Hence the Levenshtein distance in the numerator can be calculated as:
    $$ \mathcal{LD}(ref \to hyp) = \rm {cost\_of\_shortest\_path}(ref \circ \mathcal{L} \circ hyp) $$
where $\circ$ denotes FST composition.
Note that in large vocabulary continuous speech recognition (LVCSR), the static size of $\mathcal{L}$ could explode 
because the \textit{substitution} requires $O(V^2)$ space, where $V$ refers to the vocabulary size.
We follow the optimization practice in \cite{gorman2021finite} by leveraging auxiliary symbols,
so that $\mathcal{L}$ is factored as the product of two smaller FSTs.

\subsubsection{Dynamic Alternative Expansion (DAE)}
Similar to NIST's GLM machnism\footnote{https://github.com/usnistgov/SCTK/blob/master/doc/GLMRules.txt},
we enhance standard Levenshtein distance algorithm to support customizable alternative sets to deal with synonyms, contractions, abbreviations, compound words, etc on-the-fly.

\begin{figure}[h]
    \centering
    \includegraphics[width = 0.48\textwidth]{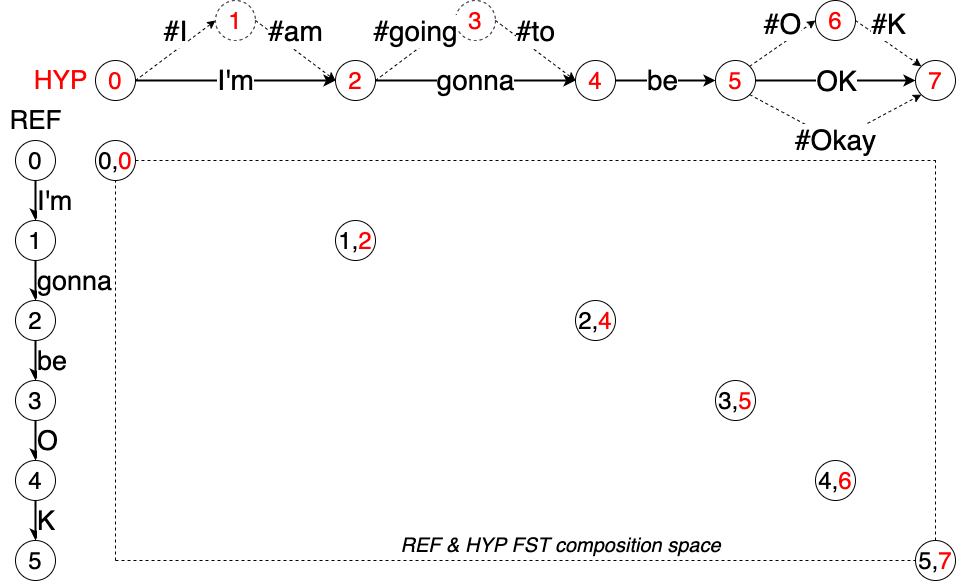}
    \caption{Levenshtein distance with Dynamic Alternative Expansion (DAE)}
    \label{fig:wfst_eval}
\end{figure}

As illustrated in Figure \ref{fig:wfst_eval}. $hyp$ is transformed from the original linear structure to a sausage-like FST, so that no matter which alternative appears in $ref$, the expanded $hyp$ can correctly match up with it. Note that we mark the expanded alternative paths with auxiliary hash tags, and Levenshtein transducer $\mathcal{L}$ is modified accordingly to disallow partial matches of these expanded segments.
Also note that dynamic alternative expansion (DAE) is only applied to the $hyp$, whereas the $ref$ remains intact. 

\begin{table*}[hb]
  \caption{Benchmark Results WER(in \%) and Ranking (quoted)}
  \label{tab:allresult_wer}
  \centering
  \scalebox{0.81}{
  \begin{tabular}{c|c|c|c|c|c|c|c|c|c|c|c}
    \toprule
    Model &\rotatebox{270}{LibriSpeech.test-clean} 
    &\rotatebox{270}{LibriSpeech.test-other}
    &\rotatebox{270}{TEDLIUM3.dev}
    &\rotatebox{270}{TEDLIUM3.test}
    &\rotatebox{270}{GigaSpeech.dev}
    &\rotatebox{270}{GigaSpeech.test}
    &\rotatebox{270}{VoxPopuli.dev}
    &\rotatebox{270}{VoxPopuli.test}
    &\rotatebox{270}{VoxPopuli.test-accented}
    &\rotatebox{270}{CommonVoice11.0.dev}
    &\rotatebox{270}{CommonVoice11.0.test}
    \\
    \midrule
aliyun\_api\_en&4.34 (12)&10.01 (12)&5.79 (12)&5.53 (12)&12.63 (11)&12.83 (11)&11.93 (9)&11.50 (9)&15.61 (5)&15.67 (9)&17.86 (9)\\
amazon\_api\_en&6.42 (16)&13.20 (15)&5.11 (8)&4.74 (9)&11.31 (8)&11.71 (8)&12.14 (11)&11.84 (11)&14.56 (4)&22.66 (15)&26.25 (16)\\
baidu\_api\_en&6.61 (17)&14.69 (16)&8.65 (14)&7.93 (14)&16.94 (13)&16.80 (13)&14.55 (15)&14.08 (15)&19.98 (11)&22.74 (16)&25.97 (15)\\
google\_api\_en&5.63 (15)&12.56 (14)&5.54 (10)&5.26 (10)&11.70 (9)&11.78 (9)&11.95 (10)&11.77 (10)&15.72 (6)&17.49 (10)&20.60 (10)\\
google\_USM\_en&2.13 (5)&4.35 (5)&3.69 (3)&3.05 (1)&8.67 (2)&9.23 (3)&9.89 (8)&9.50 (8)&14.18 (3)&8.69 (5)&10.19 (4)\\
microsoft\_sdk\_en&3.03 (10)&6.65 (10)&3.65 (2)&3.14 (2)&8.74 (3)&9.02 (1)&9.04 (5)&8.88 (5)&12.17 (1)&8.86 (6)&10.46 (5)\\
tencent\_api\_en&3.54 (11)&7.26 (11)&4.94 (7)&4.20 (5)&10.19 (6)&10.37 (6)&9.21 (6)&8.90 (6)&15.81 (7)&11.59 (7)&12.91 (7)\\

vosk\_model\_en\_large&5.17 (14)&12.36 (13)&5.74 (11)&5.41 (11)&13.84 (12)&14.08 (12)&13.22 (12)&12.52 (12)&20.27 (12)&21.66 (14)&25.46 (14)\\
deepspeech\_model\_en&6.95 (18)&21.23 (19)&17.71 (18)&18.31 (18)&32.68 (17)&31.21 (17)&28.46 (19)&29.10 (19)&38.23 (19)&43.34 (19)&47.80 (19)\\
coqui\_model\_en&4.94 (13)&15.92 (18)&17.48 (17)&18.66 (19)&34.62 (18)&32.45 (18)&27.60 (18)&27.66 (18)&37.85 (18)&35.52 (17)&40.10 (17)\\
nemo\_conformer\_ctc\_large\_en&1.95 (4)&4.26 (4)&5.11 (8)&4.69 (8)&11.70 (9)&12.01 (10)&6.68 (2)&6.61 (2)&18.24 (8)&8.44 (3)&9.23 (2)\\
nemo\_conformer\_transducer\_xlarge\_en&1.36 (1)&2.76 (1)&4.61 (5)&4.31 (6)&10.85 (7)&11.53 (7)&6.05 (1)&5.97 (1)&20.80 (14)&5.20 (1)&5.82 (1)\\
k2\_gigaspeech&2.18 (7)&5.30 (8)&3.42 (1)&3.40 (3)&8.90 (4)&9.71 (5)&9.80 (7)&9.48 (7)&12.46 (2)&14.05 (8)&17.04 (8)\\
whisper\_large\_v1&2.66 (9)&5.81 (9)&4.90 (6)&4.31 (6)&9.11 (5)&9.68 (4)&8.31 (4)&7.88 (4)&20.80 (14)&8.68 (4)&10.89 (6)\\
whisper\_large\_v2&2.14 (6)&4.65 (6)&3.70 (4)&3.55 (4)&8.53 (1)&9.19 (2)&7.82 (3)&7.17 (3)&18.83 (9)&7.88 (2)&10.01 (3)\\

data2vec\_audio\_large\_ft\_libri\_960h&1.64 (2)&3.85 (3)&8.31 (13)&7.66 (13)&18.43 (14)&17.92 (14)&13.73 (13)&13.42 (13)&19.45 (10)&18.25 (12)&21.75 (11)\\
hubert\_xlarge\_ft\_libri\_960h&1.79 (3)&3.48 (2)&8.94 (15)&8.14 (15)&18.87 (15)&18.43 (15)&14.47 (14)&13.62 (14)&20.58 (13)&18.05 (11)&21.79 (12)\\
wav2vec2\_large\_robust\_ft\_libri\_960h&2.46 (8)&5.18 (7)&9.66 (16)&9.02 (16)&19.80 (16)&18.81 (16)&14.99 (16)&14.76 (16)&21.00 (16)&18.93 (13)&22.35 (13)\\
wavlm\_base\_plus\_ft\_libri\_clean\_100h&7.00 (19)&15.68 (17)&18.75 (19)&18.21 (17)&34.78 (19)&33.02 (19)&21.55 (17)&21.16 (17)&27.94 (17)&40.17 (18)&44.96 (18)\\
    \bottomrule
  \end{tabular}
  }
\end{table*}

There are two reasons for dynamic alternative expansion instead of processing them directly in the text normalization stage. Firstly, specifying a canonical form out of these alternatives is controversial, and we as platform developers are more interested in providing DAE as a flexible mechanism rather than ironing out some specific TN (Text Normalization) policies. Secondly, with $hyp$-only expansion, the algorithm will always honor the raw form of human labels (i.e. $ref$), so that alternative sets can evolve without mutating the denominator of TER (i.e. $|ref|$), which is a desirable property for comparison purpose over time.

\section{The Benchmark}

\subsection{Results}
Table \ref{tab:allresult_wer} shows Word Error Rates (WER) and the ranking covering all models and test sets on the platform:

1) Given DeepSpeech being the SOTA back in 2014, OpenAI-Whisper model now has achieved remarkable \emph{relative WER reductions} over DeepSpeech:
78\% on LibriSpeech-test-other, 81\% on TedLium3-test, 71\% on GigaSpeech-test, 75\% on VoxPopuli-test, 50\% on VoxPopuli-accented, 79\% on CommonVoice-test.
These numbers reflect the overall progress of ASR over the last decade.

2) Open-source models tend to beat commercial services on LibriSpeech by a large margin,
which reveals some weird situations relating to the highly influential LibriSpeech dataset:
(i) The LibriSpeech benchmark may not be a preferable indicator for real-life ASR, 
therefore commercial services assign little efforts to the task in their productions.
(ii) On the other hand, open-source models apparently prioritize LibriSpeech since it is the de facto standard in ASR research,
but these systems often generalize poorly to other tasks.
The effort in pushing the LibriSpeech SOTA may turn out to be an overfitting game.
(iii) Similarly, even strong self-supervised pretrained models (such as data2vec, hubert, wav2vec and wavlm) fail to show strong generalization ability if finetuned on LibriSpeech only. 

3) As another manifestation of the scaling law, large speech models, such as Whisper, USM, and xlarge NeMo transducer,
are shown to be highly competitive even compared with the best commercial services.
More importantly, some of these large ASR systems are open-sourced.
Therefore in a foreseeable future, as these large models getting better and hardware getting cheaper,
some of the traditional ASR providers need to seek compelling reasons for customers to pay.

\subsection{Scoring subtleties}

There are various subtleties in the scoring pipeline that can affect benchmark results,
such as punctuation, interjection, non-standard word normalization, US/UK spellings, etc.
To study the importance of these subtleties,
an ablation experiment is conducted by turning off each component respectively, as shown in columns (A1-A5) from table \ref{tab:ablation2}.
\footnote{ Note that we use GigaSpeech test set for this study rather than LibriSpeech,
since GigaSpeech is curated mainly from real-world scenarios (YouTube and podcasts) on daily topics in spontaneous conversations.}


\begin{table*}[ht]
  \scriptsize
  \caption{ Effects of turning-off single component (on GigaSpeech-Test) }
  \label{tab:ablation2}
  \centering
  \scalebox{1}{
  \begin{tabular}{c|c|c|c|c|c|c|c}
    \toprule
      &  & A0 (baseline) & A1 & A2 & A3 & A4 & A5  \\
    \midrule
    \multirow{6}{2cm}{\centering Evaluation Pipeline Component (crossed = turned off) } & CASE & \Checkmark & \Checkmark  & \Checkmark & \Checkmark & \Checkmark & \Checkmark \\
    ~ & PUNC & \Checkmark & \XSolidBrush & \Checkmark & \Checkmark & \Checkmark & \Checkmark  \\
    ~ & ITJ & \Checkmark & \Checkmark & \XSolidBrush & \Checkmark & \Checkmark & \Checkmark   \\
    ~ & UK-US & \Checkmark & \Checkmark & \Checkmark  & \XSolidBrush & \Checkmark & \Checkmark  \\
    ~ & NSW & \Checkmark & \Checkmark  & \Checkmark & \Checkmark  & \XSolidBrush & \Checkmark \\
    ~ & DAE & \Checkmark & \Checkmark  & \Checkmark  & \Checkmark  & \Checkmark  & \XSolidBrush  \\
    
    \hline
    
    \midrule
    \multirow{19}{2cm}{\centering WER(\%) (Rank)} & 
microsoft\_sdk\_en & 9.02 (1) & 9.81 (1) & 9.68 (1) & 9.05 (1) & 9.02 (1) & 10.46 (1)\\
~ & whisper\_large\_v2 & 9.19 (2) & 22.43 (14) & 9.91 (3) & 9.22 (2) & 10.88 (4) & 10.57 (2)\\
~ & google\_USM\_en & 9.23 (3) & 9.85 (2) & 9.85 (2) & 9.26 (3) & 10.77 (3) & 10.75 (3)\\
~ & whisper\_large\_v1 & 9.68 (4) & 22.62 (15) & 10.39 (5) & 9.71 (4) & 11.34 (5) & 11.12 (5)\\
~ & k2\_gigaspeech & 9.71 (5) & 10.34 (3) & 10.36 (4) & 9.74 (5) & 9.72 (2) & 11.06 (4)\\
~ & tencent\_api\_en & 10.37 (6) & 20.88 (13) & 10.94 (6) & 10.39 (6) & 11.42 (6) & 11.70 (6)\\
~ & nemo\_conformer\_transducer\_xlarge\_en & 11.53 (7) & 12.35 (4) & 12.22 (7) & 11.57 (7) & 11.53 (7) & 12.78 (7)\\
~ & amazon\_api\_en & 11.71 (8) & 23.78 (16) & 12.30 (8) & 11.73 (8) & 13.30 (11) & 13.05 (8)\\
~ & google\_api\_en & 11.78 (9) & 12.41 (5) & 12.48 (9) & 11.80 (9) & 13.29 (10) & 13.32 (10)\\
~ & nemo\_conformer\_ctc\_large\_en & 12.01 (10) & 12.81 (6) & 12.62 (10) & 12.04 (10) & 12.01 (8) & 13.27 (9)\\
~ & aliyun\_api\_en & 12.83 (11) & 13.71 (7) & 13.44 (11) & 12.86 (11) & 12.83 (9) & 13.98 (11)\\
~ & vosk\_model\_en\_large & 14.08 (12) & 14.84 (8) & 14.68 (12) & 14.10 (12) & 14.08 (12) & 15.47 (12)\\
~ & baidu\_api\_en & 16.80 (13) & 17.56 (9) & 17.43 (13) & 16.83 (13) & 16.81 (13) & 17.98 (13)\\
~ & data2vec\_audio\_large\_ft\_libri\_960h & 17.92 (14) & 18.68 (10) & 18.13 (14) & 17.99 (14) & 17.92 (14) & 18.85 (14)\\
~ & hubert\_xlarge\_ft\_libri\_960h & 18.43 (15) & 19.19 (11) & 18.61 (15) & 18.50 (15) & 18.43 (15) & 19.33 (15)\\
~ & wav2vec2\_large\_robust\_ft\_libri\_960h & 18.81 (16) & 19.56 (12) & 18.93 (16) & 18.89 (16) & 18.81 (16) & 19.73 (16)\\
~ & deepspeech\_model\_en & 31.21 (17) & 31.70 (17) & 31.73 (17) & 31.24 (17) & 31.21 (17) & 32.04 (17)\\
~ & coqui\_model\_en & 32.45 (18) & 33.12 (18) & 32.72 (18) & 32.49 (18) & 32.45 (18) & 33.27 (18)\\
~ & wavlm\_base\_plus\_ft\_libri\_clean\_100h & 33.02 (19) & 33.74 (19) & 33.06 (19) & 33.07 (19) & 33.02 (19) & 33.42 (19)\\
    \bottomrule
  \end{tabular}
  }
  \begin{minipage}[c]{0.8\textwidth}
    \vspace{0.1cm}
    * Microsoft API provides a switch to disable Inverse Text Normalization(ITN), so its number in A4 (turning off NSW normalization in our pipeline) is basically the same as the baseline.
  \end{minipage}
\end{table*}

\subsubsection{PUNC}
The comparison between A1 and A0 reminds us that the processing of \emph{punctuation is vital}.
By turning off the punctuation removal component, we see a dramatic increase of WER for Whisper, tencent, amazon.
This is because different ASR systems do have quite distinctive policies for punctuation.
Therefore it requires serious and careful treatment,
otherwise the whole evaluation can be completely biased and dominated by punctuation-related errors.
Also note that proper processing of punctuation is non-trivial:
consider things like single quotes mixed with apostrophe in contraction, comma and period in numbers, hyphen in compound words, etc.

\subsubsection{ITJ}
Comparing A2 and A0, we see the processing of interjections can result in absolute WER changes around 0.5\% $\sim$ 1.0\%.
The difference seems to be mild, until we realize these numbers translate to a relative \textasciitilde 10\% WER increase.
Therefore the removal of interjections is by no means a factor that we can ignore in ASR research and evaluation on conversational ASR tasks.
The implementation is trivial though, given a list of common interjections such as 'uh', 'eh', 'um'.
\footnote{Our platform maintains common interjections at: \url{https://github.com/SpeechColab/Leaderboard/blob/master/utils/interjections_en.csv }}

\subsubsection{UK-US}
Comparing A3 and A0, we can see that the unification of US-UK spelling convention has minor effects.

\subsubsection{NSW and DAE}
Comparing A4, A5 with A0, it is apparent that NSW normalization and DAE are both crucial
\footnote{NSW component deals with the normalization of numbers, quantities, date, money, and times, etc,
and DAE module is responsible for contractions, abbreviations, and compound words.}.
Traditional ASR systems tend to generate spoken-form recognition results,
whereas commercial services and modern end-to-end systems (such as Whisper) normally yield the written-form as default.
The discrepancy between spoken and written forms is so significant,
that any serious benchmark should dive into these details to make sure the normalized hypotheses and references are consistent,
otherwise the benchmark can become pointless.
However, robust text normalization (TN) is challenging and immensely tedious due to numerous rules and long-tail exceptions.
Therefore, we highly suggest other benchmark developers to leverage and improve existing tools instead of building new TN from scratch.

\begin{figure}[h]
    \centering
    \includegraphics[width = 0.43\textwidth]{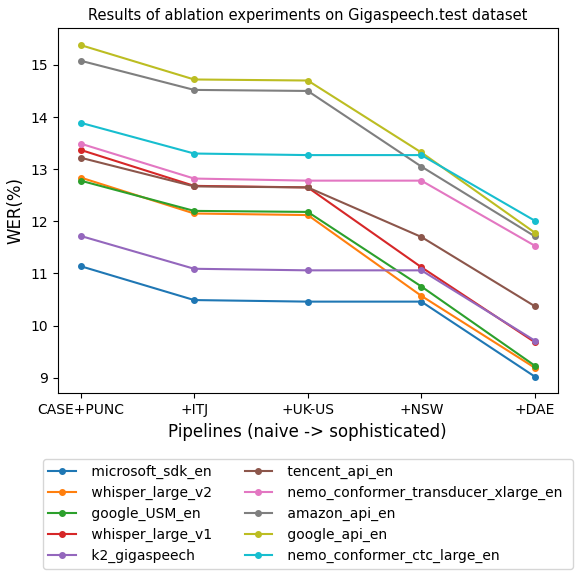}
    \caption{Results of ablation experiments on GigaSpeech.test dataset}
    \label{fig:ablation_gigaspeech}
\end{figure}

In Figure \ref{fig:ablation_gigaspeech}, we show how these individual components are stacked together. As can be seen:
i) The ranking can vary from the left (a barely working naive pipeline) to the right (a sophisticated pipeline).
ii) The sophisticated pipeline can have a relative 20\% $\sim$ 30\% WER reduction over the naive pipeline.
These observations raise a concerning revelation that technological progress can be easily shadowed by implementation details of the scoring pipeline.
For this reason, when we are discussing the absolute number of WER and the rankings of different systems,
the specific toolkit and pipeline in use should be deliberately brought into the context, to avoid misunderstanding and misinformation.

\section{Modified Token Error Rate (mTER)}
For decades, Token Error Rate (TER) has been the standard evaluation metric of Automatic Speech Recognition (ASR).
In this section, we propose a new metric to address some fundamental issues of TER. 

\subsection{Problems with TER}

\subsubsection{TER violates the metric-space axioms}
In mathematics, a \emph{metric space} needs to satisfy a small set of requirements called \emph{the metric (in)equalities}\footnote{
Consisting of following axioms:
1) positivity: $ D(x, y) \ge 0 $;
2) identity: $ D(x,y) = 0 $ if and only if $ x = y $;
3) symmetry: $ D(x,y) = D(y,x) $;
4) triangle-inequality: $ D(x,y) \le D(x, z) + D(z, y) $
}.
Obviously, TER violates the symmetry axiom, because:
$${\rm TER}(ref, hyp) \ne {\rm TER}(hyp, ref)$$

This has broad impacts in practice: 
For industry, tools and pipelines need to implement asymmetric interfaces and options, so as documentations;
For education, instructors need to emphasis the denominator calculation to their students,
as well as the relativity of insertion (INS) versus deletion (DEL).

\subsubsection{TER is ill-normalized}
TER picks $|ref|$ as the normalizer,
which makes it numerically unbounded (can easily exceed 100\%).
This becomes more confusing since it is literally called error \emph{rate}.

These problems have been identified and reported in the prior literature, and new metrics are proposed in \cite{maier2002evaluating,morris2004and}.
Besides fixing above issues, these new metrics tend to incorporate some sort of informational/cognitive measure into ASR evaluation,
which is generally the right call.
Unfortunately, a new metric needs to fight against strong inertia since the main purpose of a metric is to serve comparisons over time.
As a result, these \textit{backward-incompatible} metrics never receive widespread adoptions.

\begin{table*}[h]
  \caption{The comparison of (TER/mTER)}
  \label{tab:allresult}
  \centering
  \scalebox{0.68}{
  \begin{tabular}{c|ccccccccccc}
    \toprule
    Model &\rotatebox{270}{LibriSpeech.test-clean} 
    &\rotatebox{270}{LibriSpeech.test-other}
    &\rotatebox{270}{TEDLIUM3.dev}
    &\rotatebox{270}{TEDLIUM3.test}
    &\rotatebox{270}{GigaSpeech.dev}
    &\rotatebox{270}{GigaSpeech.test}
    &\rotatebox{270}{VoxPopuli.dev}
    &\rotatebox{270}{VoxPopuli.test}
    &\rotatebox{270}{VoxPopuli.test-accented}
    &\rotatebox{270}{CommonVoice11.0.dev}
    &\rotatebox{270}{CommonVoice11.0.test}
    \\
    \midrule
    aliyun\_api\_en & $4.34/4.33 $ & $10.01/9.99 $ & $5.79/5.79 $  & $5.53/5.53 $ & $12.63/12.63 $ & $12.83/12.83 $ & $11.93/11.93 $ & $11.50/11.49 $ & $15.61/15.61 $ & $15.67/15.51 $ & $17.86/17.67 $  \\
    amazon\_api\_en & $6.42/6.41 $ & $13.20/13.20 $ & $5.11/5.11 $  & $4.74/4.74 $ & $11.31/11.31 $ & $11.71/11.71 $ & $12.14/12.13 $ & $11.84/11.84 $ & $14.56/14.56 $ & $22.66/22.56 $ & $26.25/26.25 $  \\
    baidu\_api\_en & $6.61/6.58 $ & $14.69/14.65 $ & $8.65/8.65 $  & $7.93/7.93 $ & $16.94/16.94 $ & $16.80/16.80 $ & $14.55/14.55 $ & $14.08/14.08 $ & $19.98/19.98 $ & $22.74/22.60 $ & $25.97/25.97 $  \\
    google\_api\_en & $5.63/5.61 $ & $12.56/12.53 $ & $5.54/5.54 $  & $5.26/5.26 $ & $11.70/11.70 $ & $11.78/11.78 $ & $11.95/11.93 $ & $11.77/11.76 $ & $15.72/15.68 $ & $17.49/17.31 $ & $20.60/20.58 $  \\
    google\_USM\_en & $2.13/2.13 $ & $4.35/4.35 $ & $3.69/3.69 $  & $3.05/3.05 $ & $8.67/8.60 $ & $9.23/9.23 $ & $9.89/9.89 $ & $9.50/9.50 $ & $14.18/14.18 $ & $8.69/8.69 $ & $10.19/10.19 $  \\
    microsoft\_sdk\_en & $3.03/3.02 $ & $6.65/6.64 $ & $3.65/3.65 $  & $3.14/3.14 $ & $8.74/8.69 $ & $9.02/9.02 $ & $9.04/9.00 $ & $8.88/8.84 $ & $12.17/12.09 $ & $8.86/8.84 $ & $10.46/10.46 $   \\
    tencent\_api\_en & $3.54/3.54 $ & $7.26/7.26 $ & $4.94/4.94 $  & $4.20/4.20 $ & $10.19/10.19 $ & $10.37/10.37 $ & $9.21/9.21 $ & $8.90/8.90 $ & $15.81/15.81 $ & $11.59/11.52 $ & $12.91/12.91 $ \\
    \midrule
    vosk\_model\_en\_large& $5.17/5.16 $ & $12.36/12.36 $ & $5.74/5.74 $  & $5.41/5.41 $ & $13.84/13.84 $ & $14.08/14.08 $ & $13.22/13.08 $ & $12.52/12.41 $ & $20.27/19.61 $ & $21.66/21.29 $ & $25.46/25.42 $   \\
    deepspeech\_model\_en & $6.95/6.95 $ & $21.23/21.23 $ & $17.71/17.71 $  & $18.31/18.31 $ & $32.68/32.68 $ & $31.21/31.21 $ & $28.46/28.46 $ & $29.10/29.10 $ & $38.23/38.23 $ & $43.34/43.34 $ & $47.80/47.80 $  \\
    coqui\_model\_en      & $4.94/4.93 $ & $15.92/15.92 $ & $17.48/17.48 $  & $18.66/18.66 $ & $34.62/34.62 $ & $32.45/32.45 $ & $27.60/27.60 $ & $27.66/27.66 $ & $37.85/37.85 $ & $35.52/35.52 $ & $40.10/40.10 $  \\
    nemo\_conformer\_ctc\_large\_en      & $1.95/1.95 $ & $4.26/4.26 $ & $5.11/5.10 $  & $4.69/4.69 $ & $11.70/11.68 $ & $12.01/12.01 $ & $6.68/6.68 $ & $6.61/6.61 $ & $18.24/18.24 $ & $8.44/8.44 $ & $9.23/9.23 $  \\
    nemo\_conformer\_transducer\_xlarge\_en      & $1.36/1.36 $ & $2.76/2.76 $ & $4.61/4.61 $  & $4.31/4.31 $ & $10.85/10.85 $ & $11.53/11.53 $ & $6.05/6.05 $ & $5.97/5.97 $ &  $20.80/20.80 $ & $5.20/5.20 $ & $5.82/5.82 $  \\
    k2\_gigaspeech      & $2.18/2.18 $ & $5.30/5.30 $ & $3.42/3.42 $  & $3.40/3.40 $ & $8.90/8.90 $ & $9.71/9.71 $ & $9.80/9.80 $ & $9.48/9.48 $ &  $12.46/12.46 $ & $ 14.05/14.05$ & $17.04/17.04 $  \\
    whisper\_large\_v1      & $2.66/2.66 $ & $5.81/5.80 $ & $4.90/4.90 $  & $4.31/4.31 $ & $9.11/9.11 $ & $9.68/9.68 $ & $8.31/8.31 $ & $7.88/7.87 $ & $20.80/20.80 $ & $8.68/8.64 $ & $10.89/10.82 $ \\
    whisper\_large\_v2      & $2.14/2.14 $ & $4.65/4.65 $ & $3.70/3.70 $  & $3.55/3.55 $ & $8.53/8.53 $ & $9.19/9.19 $ & $7.82/7.82 $ & $7.17/7.17 $ & $18.83/18.83 $ & $7.88/7.86 $ & $10.01/9.96 $ \\
    \midrule
    data2vec\_audio\_large\_ft\_libri\_960h      & $1.64/1.64 $ & $3.85/3.85 $ & $8.31/8.26 $  & $7.66/7.66 $ & $18.43/18.27 $ & $17.92/17.92 $ & $13.73/13.66 $ & $13.42/13.34 $ & $19.45/19.07 $ & $18.25/18.17 $ & $21.75/21.63 $   \\
    hubert\_xlarge\_ft\_libri\_960h      & $1.79/1.79 $ & $3.48/3.48 $ & $8.94/8.85 $  & $8.14/8.14 $ & $18.87/18.70 $ & $18.43/18.43 $ & $14.47/14.39 $ & $13.62/13.53 $ & $20.58/20.03 $ & $18.05/17.95 $ & $21.79/21.63 $  \\
    wav2vec2\_large\_robust\_ft\_libri\_960h     & $2.46/2.46 $ & $5.18/5.18 $ & $9.66/9.52 $  & $9.02/8.97 $ & $19.80/19.45 $ & $18.81/18.75 $ & $14.99/14.83 $ & $14.76/14.60 $ & $21.00/20.19 $ & $18.93/18.86 $ & $22.35/22.22 $  \\
    wavlm\_base\_plus\_ft\_libri\_clean\_100h      & $7.00/7.00 $ & $15.68/15.68 $ & $18.75/18.64 $  & $18.21/18.21 $ & $34.78/34.78 $ & $33.02/33.02 $ & $21.55/21.48 $ & $21.16/21.07 $ & $27.94/27.24 $ & $40.17/40.17 $ & $44.96/44.96 $   \\
    \bottomrule
  \end{tabular}
  }
\end{table*}

\begin{figure*}[htb]
\centering
\caption{An example showing the difference in values between mTER and TER.}
\scriptsize
\begin{verbatim}
----------------------------------------------------- an example -----------------------------------------------------
{"uid":YOU1000000117_S0000168, "TER":76.92, "mTER":43.48, "cor":13,"sub":0, "ins":10, "del":0}
  REF  : FOR OLDER KIDS THAT CAN BE THE SAME *   WE DO IT AS ADULTS *   *    *           *     *   *   *    *   *   
  HYP  : FOR OLDER KIDS THAT CAN BE THE SAME WAY WE DO IT AS ADULTS FOR MORE INFORMATION VISIT WWW DOT FEMA DOT GOV
  EDIT :                                     I                      I   I    I           I     I   I   I    I   I 
----------------------------------------------------------------------------------------------------------------------
\end{verbatim}
\label{fig:example-mter}
\end{figure*}

\subsection{mTER}
In \cite{Vitányi2009}, Paul M. B. Vitányi et al proposed a universal similarity metric based on Kolmogorov complexity,
called Normalized Information Distance (NID):
\begin{equation}
\begin{split}
{\rm NID}(x, y) &\doteq \frac{ \max \{ K(x|y), K(y|x) \} }{ \max \{ K(x), K(y) \} } \\
&= \frac{\max \{ K(x|y), K(y|x) \} } {\max{ \{ K(x|\epsilon), K(y|\epsilon) \} } }
\end{split}
\end{equation}
Due to the incomputability of Kolmogorov complexity (subjected to Turing's halting problem), NID can only serve as a theoretical framework.
Some practical approximations of NID, such as Normalized Compression Distance (NCD)
\cite{cilibrasi2005clustering},
Normalized Web Distance (NWD) have been investigated for wide range of tasks
including genome phylogeny \cite{li2001information}, plagiarism detection \cite{chen2004shared} and information retrieval \cite{cilibrasi2009normalized}, etc.

The similarity between Conditional Kolmogorov complexity and Levenshtein distance is striking.
Since $K(O|I)$ depicts the theoretical \emph{minimal program length} to generate output $O$ given input $I$ (both encoded in bit-sequences) within a Universal Turing Machine,
whereas $\mathcal{LD}(I \to O)$ takes a more pragmatic approach by counting the \emph{minimal edit instructions} required to transform $I$ into $O$ (both encoded as token sequences) within a heavily constrained instruction set \{INS, DEL, SUB\}.
In the scope of this paper, it is not necessary to establish more formal connection between $K$ and $\mathcal{LD}$ though
\footnote{
    Actually, the algorithmic information framework can provide deeper insights.
    For instance: The ``INS" and ``DEL" instructions should by no means be equally weighted from an information perspective,
    because it requires immensely more complexity for DaVinci to \emph{create} the ``Mona Lisa" than someone to \emph{erase} it, i.e.:
    $K(MonoLisa|\epsilon) \gg K(\epsilon|MonoLisa) $.
    So in principle, more cognitive-aware metrics can be designed following the algorithmic information theory.
},
we denote it as a simple heuristic:
\begin{equation}
    K(O|I) \Leftrightarrow \mathcal{LD}(I \to O)
\end{equation}

Given this heuristic, now observing the definition of NID (especially pay attention to the form of the denominator as the metric normalizer),
we propose modified Token Error Rate:
\begin{equation}
\begin{split}
    &{\rm mTER}(ref, hyp) \\
    &\doteq \frac{\max \{ \mathcal{LD}(hyp \to ref), \mathcal{LD}(ref \to hyp) \} } {\max \{ \mathcal{LD}(\epsilon \to ref),  \mathcal{LD}(\epsilon \to hyp) \} } \\
    &= \frac{\mathcal{LD}(ref \to hyp)}{\max \{ |ref|,  |hyp| \} }
\end{split}
\end{equation}
Note that the above numerator simplification is based on the fact that $\mathcal{LD}$ is symmetric with regard to $ref$ and $hyp$.

Compared to the vanilla TER in equation (\ref{eq:vanilla-TER}), it is merely a modification to the denominator:
$|ref| \Rightarrow \max \{|ref|, |hyp|\}$.
However, the resulting mTER is symmetric, gracefully normalized and bounded between $[0\%, 100\%]$, and perfectly conforms to the aforementioned metric-space axioms.

\subsection{Empirical study on mTER and TER}

\subsubsection{Dataset-Level comparison}
Table \ref{tab:allresult} compares TER and mTER on all models and test sets on the platform. About 66\% of the mTER numbers are exact same as TER. And for the rest, WER discrepancies are minor (\textless4\% relatively). That basically means mTER is backward-compatible, researchers will still be able to compare with TER from prior works.

\subsubsection{Utterence-Level comparison}

\begin{figure}[h]
    \centering
    \includegraphics[width = 0.47\textwidth]{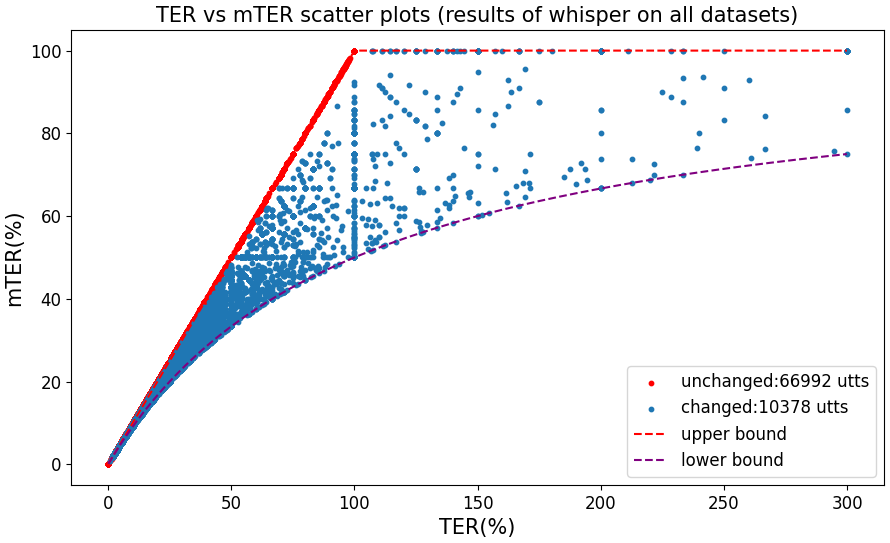}
    \caption{TER vs mTER scatter plots}
    \label{fig:scatter}
\end{figure}

Figure \ref{fig:scatter} shows mTER vs TER of all testing utterances on \emph{whisper\_large\_v1} model.
The samples on the red diagonal line represent utterances whose mTER is precisely consistent with TER, accounting to 87\% of the total samples.
The remaining 13\% samples illustrated how mTER and TER are correlated,
and how mTER wrapped overflowed TER (i.e. TER $>$ 100\%) into [0\%, 100\%].

Figure \ref{fig:example-mter} provides an utterance from our benchmark on Whisper,
showing a closer look at the discrepancy between TER and mTER.
The recognition result contains 23 words in total, among which 9 words are hallucinated.
In this case, mTER (43.38\%) is intuitively better than TER (76.92\%) because it reflects the valid portion of the ASR output more precisely.
TER is vulnerable to overflow, especially in short utterances and for insertion errors like model hallucination.

To sum up, through large-scale empirical experiments, we found the proposed mTER is robust.
i) It is no longer necessary for scoring tools, pipelines, and documentation to differentiate $ref$ and $hyp$, since mTER is symmetric.
ii) The numbers of evaluation results are guaranteed to be normalized, with no more confusing \emph{overflowed error rate}.
In practice, the adoption of mTER is demonstrated to be backward-compatible the with the existing TER in dataset level evaluations.


%

\section{Conclusion}
This paper introduces SpeechColab Leaderboard, an open-source evaluation platform for Automatic Speech Recognition (ASR). Base on the proposed platform, we conduct and report an extensive benchmark, revealing the landscape of state-of-the-art ASR systems from both research and industry. In addition, our study quantitatively investigates the relevance of different components within the evaluation pipeline, which presents a valuable reference for the community to build serious ASR benchmarks in the future. 
Furthermore, by leveraging our platform and the benchmark results, we propose a new metric, modified Token Error Rate (mTER), which is more robust and elegant than the traditional Token Error Rate (TER).
In the future, we intend to incorporate more datasets and models into the platform.

\section*{Acknowledgments}
We would like to thank Yong LIU for his help to integrate the TEDLIUM-3 evaluation sets.

\bibliographystyle{IEEEtran}
\bibliography{mybib}

 





\vfill

\end{document}